\documentclass[twocolumn]{article}
\usepackage[margin=1.5cm]{geometry}
\usepackage{amsmath,amsfonts}
\usepackage{graphicx}
\usepackage{textcomp}
\usepackage{xcolor}
\usepackage{subcaption}
\usepackage{enumitem}
\usepackage{comment}
\usepackage{multirow}
\usepackage{url}
\usepackage{algorithm}
\usepackage{algpseudocode}
\usepackage{algorithmicx}
\usepackage{cuted}
\usepackage{authblk}
\usepackage[square,numbers]{natbib}
\usepackage[hidelinks]{hyperref}

\usepackage{hyperref}
\newcommand{\keywords}[1]{%
  \vspace{0.5em}%
  \noindent\textbf{Keywords:} #1%
}

\date{}

\begin{document}

\title{EmeraldMind: A Knowledge Graph–Augmented Framework for Greenwashing Detection}

\author[1]{Georgios Kaoukis\textsuperscript{*}}
\author[1]{Ioannis Aris Koufopoulos\textsuperscript{*}}
\author[1]{Eleni Psaroudaki\textsuperscript{\dag}}
\author[1]{Danae Pla Karidi\textsuperscript{\dag}}
\author[1,3]{\\Evaggelia Pitoura}
\author[2]{George Papastefanatos}
\author[1,3]{Panayiotis Tsaparas}
\affil[1]{Archimedes, Athena Research Center, Greece}
\affil[2]{IMSI, Athena Research Center, Greece}
\affil[3]{University of Ioannina, Greece}

\maketitle
  \begingroup
  \renewcommand\thefootnote{\fnsymbol{footnote}}
  \footnotetext[1]{These authors contributed equally to this work.}
  \footnotetext[2]{Corresponding authors: \{\href{mailto:h.psaroudaki@athenarc.gr}{h.psaroudaki},%
\href{mailto:danae@athenarc.gr}{danae}\}@athenarc.gr}
  \endgroup

\begin{abstract}
As AI and web agents become pervasive in decision-making, it is critical to design intelligent systems that not only support sustainability efforts but also guard against misinformation. Greenwashing, i.e., misleading corporate sustainability claims, poses a major challenge to environmental progress. To address this challenge, we introduce EmeraldMind, a fact-centric framework integrating a domain-specific knowledge graph with retrieval-augmented generation to automate greenwashing detection. EmeraldMind builds the EmeraldGraph from diverse corporate ESG (environmental, social, and governance) reports, surfacing verifiable evidence, often missing in generic knowledge bases, and supporting large language models in claim assessment. The framework delivers justification-centric classifications, presenting transparent, evidence-backed verdicts and abstaining responsibly when claims cannot be verified. Experiments on a new greenwashing claims dataset demonstrate that EmeraldMind achieves competitive accuracy, greater coverage, and superior explanation quality compared to generic LLMs, without the need for fine-tuning or retraining.
\end{abstract}


\section{Introduction}
\label{sec:intro}

Greenwashing refers to the corporate practice of conveying a misleading impression of environmental responsibility through advertisements, statements across media channels, and traditional communication platforms. Common examples include presenting regulatory compliance as a product's environmental benefit, or advertising green claims to entire products when they pertain only to specific parts or aspects.
Greenwashing misleads consumers and investors with false sustainability claims, undermining trust, hindering genuine environmental efforts, and allowing harmful practices to persist while delaying urgent action on climate change.

Journalists assess potential greenwashing by seeking evidence in regulatory records from authorities such as the Advertising Standards Authority (ASA) or corporate Environmental, Social, and Governance (ESG) reports. ESG reports are documents that organizations publish on an annual basis to disclose their performance and practices, via standardized, non-financial metrics, i.e., key performance indicators (KPIs). These KPIs follow frameworks defined by the European Union's Sustainable Finance Disclosure Regulation (SFDR) and Corporate Sustainability Reporting Directive (CSRD). Monitoring these metrics and cross-checking them against other corporate information, such as advertising campaigns and news coverage,  can uncover inconsistencies, including environmental impacts that are inaccurately reported or overstated. 

Greenwashing detection represents a specialized subset of fact-checking research; however, it presents some unique challenges when realized through automated retrieval-augmented generation (RAG) pipelines. The first refers to the extraction and modeling of domain-specific information that can be used in automated verification pipelines \cite{gupta2024knowledge,bronzini2024glitter,kiepura-lam-2025-climatecheck2025}. 
Traditional fact-checking pipelines relying on generic sources, such as scholarly literature or general web information, often fail to retrieve and assess the specialized evidence needed for metrics like emissions KPIs, which may also be proprietary. As a result, the verification results can be incomplete or misleading \cite{chillrud2021evidence,leippold2025automated}.  Thus, \textit{greenwashing detection demands tailored retrieval and processing of multimodal content from specialized knowledge sources like ESG reports and regulatory records beyond general repositories}.

Second, the design space of RAG pipelines can be vast, with systems sourcing and combining information from diverse repositories (e.g., internal knowledge bases, web data, ESG reports). This leads to highly variable outputs that depend heavily on retrieval quality and prompting strategies and sensitivities that are further amplified by the ambiguity inherent in the definition of greenwashing \cite{calamai2025corporate}. In high-stakes domains like sustainability, a simple true/false label is insufficient;  \textit{models must provide evidence-backed justifications to ensure trust and accountability among stakeholders}
\cite{zeng2024justilm,rahman2025hallucination,leippold2025automated,qazi2025scaling}. Accuracy-focused evaluation can be misleading, as high scores may result from excessive abstentions, i.e., cases where the system does not make a judgment, or ungrounded guesses, both of which obscure the system’s reasoning and limit practical adoption. For real-world use, RAG-based systems must maintain high coverage, balancing accuracy with comprehensive, evidence-backed assessments. In this respect, the challenge lies in designing a suitable RAG pipeline that \textit{maximizes the number of claims assessed with well-supported, evidence-based justifications, while minimizing both the unsupported predictions and abstentions.}

Third, \textit{a major obstacle for training, fine-tuning, and evaluating greenwashing detection pipelines remains the scarcity of annotated datasets} \cite{calamai2025corporate}. While domain-specific fine-tuning yields superior performance \cite{li2024self},  it requires substantial annotated data, which is costly, time-intensive to create, and demands significant domain expertise.
Overall, the greenwashing detection problem can be defined as: \textit{\textbf{Given a textual sustainability claim, determine whether it constitutes greenwashing and provide a fact-based justification; if the evidence is insufficient, abstain from issuing a verdict}}. 

In this paper, we propose \texttt{EmeraldMind}, which is a domain-specific RAG framework for greenwashing detection. Our framework is built on two complementary custom sustainability knowledge stores: the \emph{EmeraldGraph} that captures ESG-specific entities and relations, as extracted from ESG reports, KPI definitions \cite{EFFAS_DVFA2009}, and widely known greenwashing claim examples, and the \emph{EmeraldDB} that captures raw ESG text, with tailored retrieval mechanisms to gather relevant evidence for each claim. By grounding LLM reasoning in this domain-specific knowledge, our system produces a verdict plus an evidence-backed justification, or abstains when insufficient facts exist, enabling auditable outputs that address accuracy-only evaluation limitations. To evaluate our framework, we propose a semi-synthetic data benchmark named \texttt{EmeraldData}, which can also be used to fine-tune other models. To the best of our knowledge, this is the first work to propose an end-to-end \emph{knowledge-based retrieval-augmented generation} framework for \emph{greenwashing detection}.
Our key contributions can be summarized as follows. 
\begin{enumerate} [wide]
\item \textbf{\texttt{EmeraldMind} framework}: 
We introduce a domain-specific, knowledge-based, and abstention-aware RAG pipeline. Given a sustainability claim, \texttt{EmeraldMind} produces a verdict (greenwashing, not greenwashing, or abstain) along with a natural-language justification grounded in retrieved evidence.
\item \textbf{\emph{EmeraldGraph} knowledge graph:} We develop a structured sustainability knowledge graph centered on company-specific ESG entities and relations. Our graph retrieval algorithm prioritizes critical entities such as companies and KPIs to ensure that relevant context is extracted for responsible reasoning and precise claim evaluation.
\item \textbf{\texttt{EmeraldData} benchmark:} We release a novel evaluation dataset for greenwashing detection, comprising 620 semi-synthetic corporate sustainability claims, each paired with a ground-truth label. This benchmark facilitates transparent evaluation and comparative testing of greenwashing detection systems.
\item \textbf{Experimental evaluation:} We conduct a thorough experimental evaluation, comparing \texttt{EmeraldMind}'s variants, demonstrating superior performance and emphasizing the importance of the critical role of justification quality in greenwashing detection evaluation.
\end{enumerate}

The paper structure is: Section~\ref{sec:related} reviews related work, Section~\ref{sec:emeraldMind} presents an overview of the \texttt{EmeraldMind} framework; Section~\ref{sec:stores} details the construction of the evidence stores; Section~\ref{sec:inference} describes the retrieval and reasoning under different variants;
Section~\ref{sec:datasets} introduces the \texttt{EmeraldData} benchmark; Section~\ref{sec:exp} reports quantitative experimental results and justification-quality analysis; finally, Section~\ref{sec:concl} concludes and outlines future work.

\section{Related Work}
\label{sec:related}

\paragraph{Fact-Checking and Greenwashing Detection with LLMs.}
Automated fact-checking systems often rely on structured resources to ground verification and produce explanations. 
While large language models (LLMs) demonstrate promising capabilities, their fact-checking accuracy is still inferior to human experts \cite{caramancion2023news}. Specialized models fine-tuned on domain-specific data yield the best performance~\cite{li2024self}; however, their success hinges on the availability of substantial annotated datasets, the creation of which is costly and time-consuming to produce, demanding significant domain expertise.

To address the lack of labeled data, both supervised and unsupervised methods have been proposed to generate synthetic fact-checking training examples \cite{wright2022generating,meng2022generating}. The UNOWN framework \cite{bussotti2024unknown} is a notable advancement, combining textual and tabular modalities to automatically create fact-checking datasets at scale,  improving scalability and reducing annotation overhead.

In the environmental communication domain, greenwashing detection presents additional challenges due to ambiguous definitions, context-dependent claims, and scarcity in the datasets \cite{calamai2025corporate}. Recent work has explored the use of machine learning and LLMs to evaluate ESG-related statements \cite{calamai2025corporate}. For example, the \texttt{ESGenius} benchmark provides structured reasoning datasets for evaluating environmental question-answering performance of LLMs \cite{esgenius2025}, while Bronzini et al. \cite{bronzini2024glitter} leverage RAG methods for analyzing disclosure similarity across companies and their relation to ESG ratings. These efforts underline the potential of LLM-driven reasoning to identify deceptive sustainability narratives, but they lack explicit grounding in a structured ESG knowledge context.

To the best of our knowledge, the only available dataset for greenwashing detection is the \texttt{GreenClaims} \cite{greenwashingdataset}, introduced by Fornasiero \cite{fornasieroexploring} and contains third-party verified cases derived from reputable news sources and regulatory records like ASA records. However, its small size limits broader applicability,  highlighting the ongoing need for expanded and continuously updated datasets \cite{calamai2025corporate}. 

\paragraph{Knowledge Graphs and RAG for ESG Fact-Checking.}
The hallucination problem in LLM outputs motivated the integration of retrieval-augmented generation (RAG) methods, which incorporate relevant external evidence into LLM prompts to improve factual accuracy \cite{rahman2025hallucination}. By retrieving relevant evidence before generation, RAG systems strengthen verification-oriented tasks. Examples from the Environmental domain include ChatReport \cite{ni2023chatreport}, and ChatClimate \cite{vaghefi2023chatclimate}, which incorporate ESG-specific retrieval and summarization for sustainability news and corporate reports. These pipelines, however, primarily focus on open-ended question answering rather than binary greenwashing claim verification.

Building on these foundations, Knowledge-graph (KG)-based RAG solutions integrate entity-level reasoning into retrieval and generation. Empirical studies show that graph-based augmentation improves factual precision and interpretability over generic text retrieval approaches.  \cite{gupta2024knowledge} for instance, uses RAG for question answering from news articles. However, none of the ESG-domain-specific existing methods provide an end-to-end pipeline for greenwashing detection in sustainability claims.

In the ESG domains, several structured ontologies and knowledge graphs have been developed. OntoSustain \cite{zhou2023ontosustain,usmanova2024structuring} formalizes corporate sustainability concepts around Global Reporting Initiative (GRI) indicators, while KnowUREnvironment \cite{islam2022knowurenvironment} and SustainGraph \cite{fotopoulou2022sustaingraph} integrate environmental and Sustainable Development Goals related entities into comprehensive graphs. Other efforts, such as RSOKG \cite{zhou2024towards} and ESGOnt \cite{vijaya2025esgont}, unify standards across the GRI and the European Sustainability Reporting Standards (ESRS), reinforcing the value of structured sustainability indicators. However, despite their value, these ontologies have not yet been incorporated into end-to-end greenwashing detection pipelines.
Automatic and semi-structured KG construction remains an active research direction \cite{zhong2023comprehensive,chen2023autokg}. 

\section{The EmeraldMind Framework}
\label{sec:emeraldMind}
In this section, we present an overview of the \texttt{EmeraldMind} framework. \texttt{EmeraldMind} is a domain-specific knowledge-graph RAG-based framework for sustainability claim verification and greenwashing detection. Given an input claim, \texttt{EmeraldMind} assesses its veracity and decides whether it constitutes greenwashing. However, such claims are often vague or self-declarative and cannot be reliably validated using the language model’s internal knowledge alone. Consequently, \texttt{EmeraldMind} is designed to produce verdicts that are both evidence-based and explainable, grounding its decisions in an external ESG context rather than relying solely on the LLM’s latent knowledge.

\begin{figure}[ht!]
    \centering
    \includegraphics[width=0.7\linewidth]{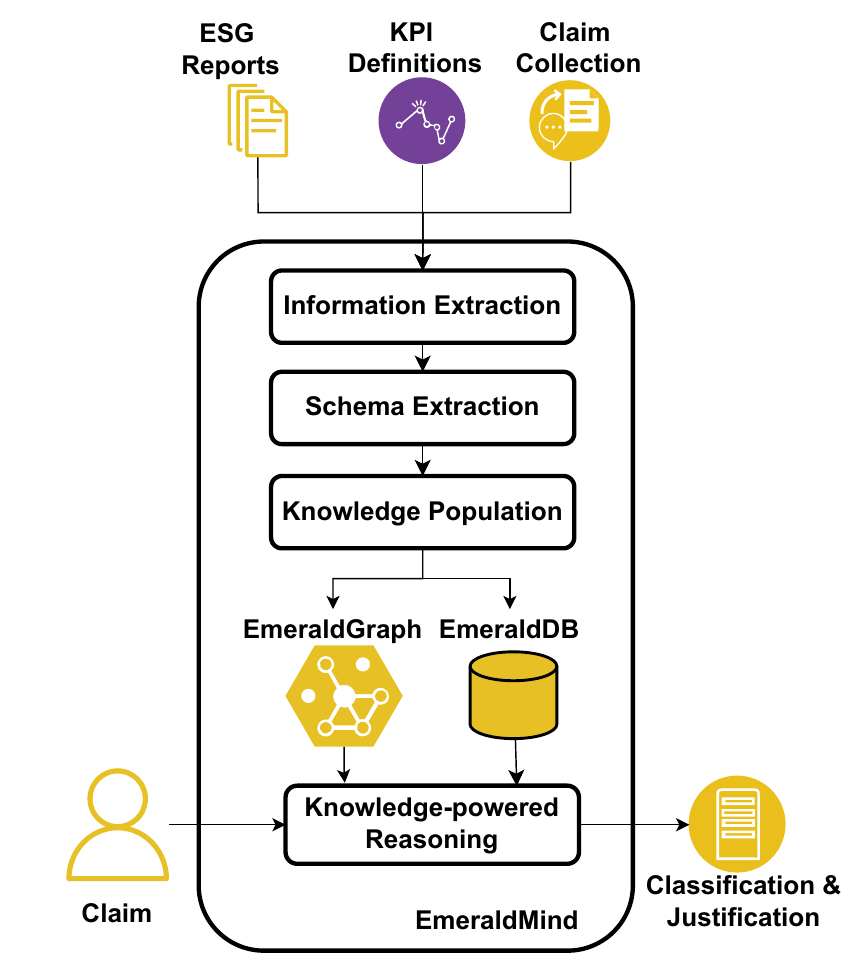}
    \caption{\texttt{EmeraldMind} Framework}
    \label{fig:framework}
\end{figure}

To this end, \texttt{EmeraldMind} adopts a two-stage architecture that explicitly separates the construction of evidence from its use in reasoning (see Figure~\ref{fig:framework}). The \textbf{Evidence Stores construction phase} (presented in Section~\ref{sec:stores}) serves a critical role: it transforms unstructured multimodal ESG reports into both structured, queryable representations and vectorized natural language chunks.
This phase normalizes heterogeneous content, induces a semi-automatically derived KPI sub-schema, and constructs two complementary stores: \textit{EmeraldGraph}, a property graph encoding ESG entities and relations, and \textit{EmeraldDB}, a document repository that preserves textual context and provenance metadata. Together, they constitute the verifiable \texttt{EmeraldMind} Evidence Stores. 

During the \textbf{Knowledge-powered Reasoning phase} (presented in Section~\ref{sec:inference}), the framework receives a user-provided claim and operates over the preconstructed evidence stores. By grounding the claim in \textit{EmeraldGraph} and retrieving semantically related chunks from \textit{EmeraldDB}, we create the context to support the reasoning. 
The final classification module delivers a verdict, classifying the claim as greenwashing or not, or abstains if the evidence is inconclusive. In all cases, the system generates a concise justification grounded in the retrieved facts. 

\section{EmeraldMind Evidence Stores}
\label{sec:stores}
In this section, we detail the construction of the evidence stores, which comprise the \emph{EmeraldDB} document store and the \emph{EmeraldGraph} knowledge graph. Given ESG reports, KPI definitions, and claim collections, the pipeline executes the following stages: (1) Information Extraction, which extracts parsed text from the reports, (2) Schema Extraction, which induces a domain schema $T$ for ESG concepts, (3) Knowledge Population that populates the evidence stores into the complementary architectures of the \emph{EmeraldDB} and the \emph{EmeraldGraph}.

\paragraph{EmeraldDB.}
\emph{EmeraldDB} is a vectorized document store that helps retrieve relevant information to support reasoning. Each ESG report is broken down into smaller sections, and key metadata for these sections, such as the report name, the referenced company, year, and page, are stored in \emph{EmeraldDB} so they can be efficiently accessed when needed.

\paragraph{EmeraldGraph.}
\emph{EmeraldGraph} is a domain-specific knowledge graph. It offers the following features: (1) it captures ESG data (e.g., facility-level emissions) that are typically absent from generic knowledge graphs, (2) it enforces structural clarity by distinguishing semantically similar but logically distinct entities (e.g., targets vs.\ actual performance), and (3) it improves auditability by grounding verdicts in explicit reasoning paths (e.g., \texttt{Company} $\to$ \texttt{reportsKPI}) rather than opaque LLM justifications.
\emph{EmeraldGraph} is defined as a labeled property graph $G=(V,E,T,S,\mathcal{A},\tau,L,p)$, where:
\begin{itemize}
    \item $V$ is the set of nodes, each representing a real-world entity mentioned in an ESG report (e.g., a company, facility, KPI observation, sustainability goal, etc.).
    \item $T$ is a finite set of entity types (node labels) used in the graph schema (e.g., \texttt{Company},\texttt{Facility},\texttt{KPIObservation}, etc.).
    \item $\tau: V \to T$ is the function associating every node $v\in V$ with its entity type $\tau(v)\in T$. For example, $\tau(v)$ = Facility if $v$ represents a plant.
    \item $L$ is the set of relationship types (edge labels).
    \item $E \subseteq V \times L \times V$ is the set of directed labeled edges $(u, l, v)$ representing  facts, e.g., (\texttt{ACME}, \texttt{reportsKPI},   \texttt{EmissionObservation}). 
    \item $S \subseteq T \times L \times T$ is the domain schema, i.e., the set of allowable typed relationships. For example, $(Facility, locatedIn, Location) \in S$ constrains the graph construction. 
    \item $\mathcal{A}$ is the set of key--value (properties) pairs. For example, \texttt{KPIObservation}: \{\texttt{value}, \texttt{unit}, \texttt{year}\}).
    \item $p: V \cup E \to \mathcal{A}$ maps each node/edge to its properties (e.g., $ p(v) = \{value:2300, unit:tCO2e, year:2025\}$).
\end{itemize}

The construction of $G$ requires (1) assigning each node $v$ a type $\tau(v)\in T$ and (2) attaching its properties $p(v)$. A central challenge is schema consistency. Each fact $(u,\ell,v)$ must be allowed by $S$ (i.e., $\tau(u)\xrightarrow{\ell}\tau(v)$), and each entry in $p$ must satisfy the type constraints (e.g., numeric fields for KPIObservation nodes).

\paragraph{Information Extraction.} It processes raw ESG reports into parsed documents used by \emph{EmeraldGraph} and \emph{EmeraldDB}. Corporate ESG reports are highly multimodal: narrative text, tables, charts, and figures all carry quantitative information. The main challenge is to extract the full semantic content without losing context \cite{bronzini2024glitter,zou2024esgreveal}. 

\begin{figure}[ht!]
    \centering
\includegraphics[width=0.9\linewidth]{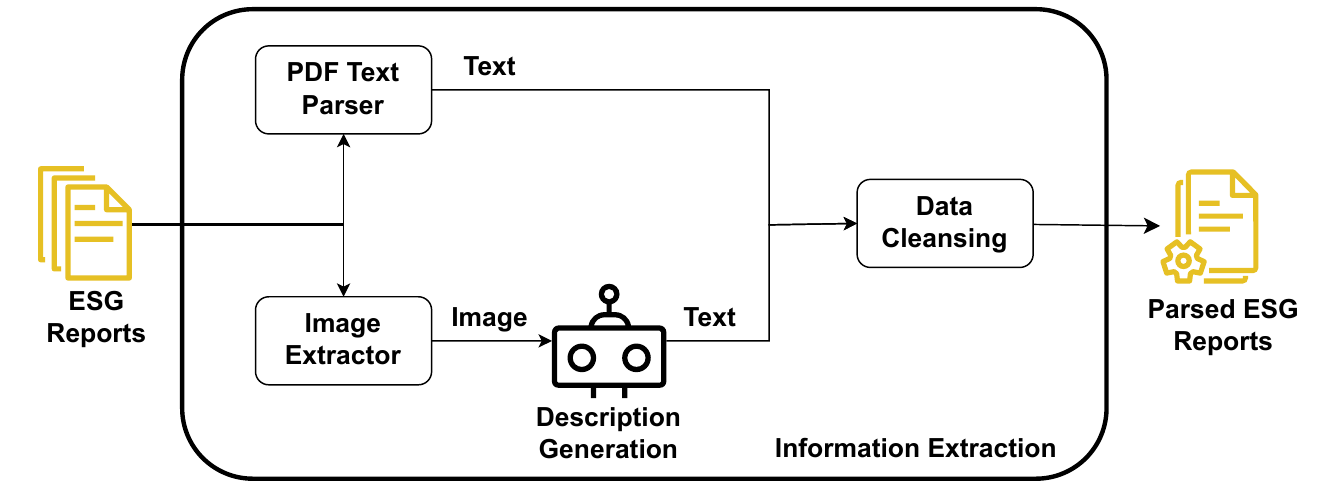}
    \caption{Information Extraction}
    \label{fig:infoExtraction}
\end{figure}

We employ a two-channel information extraction pipeline that processes textual and multimodal content separately, as shown in Figure~\ref{fig:infoExtraction}. The text channel uses a PDF parser (PyMuPDF \cite{pymupdf}) to extract paragraphs, headings, footnotes, metadata, and tables. The multimodal channel renders each page as an image and uses a vision-language model to recover semantic content from charts, figures, and tables, including structured fields (series, axes, units), captions, and descriptive context that would otherwise be lost. We reconcile outputs from both channels through alignment and deduplication, retaining both text and structured representations.

For each report, we produce a parsed representation consisting of text spans, normalized tables, and chart or figure descriptions with associated metadata. \emph{EmeraldDB} stores these spans as retrieval passages, forming a document store for semantic retrieval. \emph{EmeraldGraph} consumes the same parsed structures as input to downstream schema extraction and knowledge population, where typed entities and relations are constructed.

\paragraph{Schema Extraction.} It synthesizes a domain-specific schema $S$ that constrains all entities and relations in \emph{EmeraldGraph}. $S$ defines the allowed entity types, relation labels, and attribute domains. For example, it ensures that a node of type \texttt{Facility} can only be linked to a node of type \texttt{Location} via a \texttt{locatedIn} relation.

ESG reports are structurally heterogeneous and semantically inconsistent. The same KPI may use different labels, units, and table layouts, and claims often mix targets, baselines, and actuals. The main challenge is to design a schema that is expressive enough for diverse ESG disclosures but restrictive enough to avoid invalid entities and relations. 

\begin{figure}[h]
    \centering
    \includegraphics[width=0.9\linewidth]{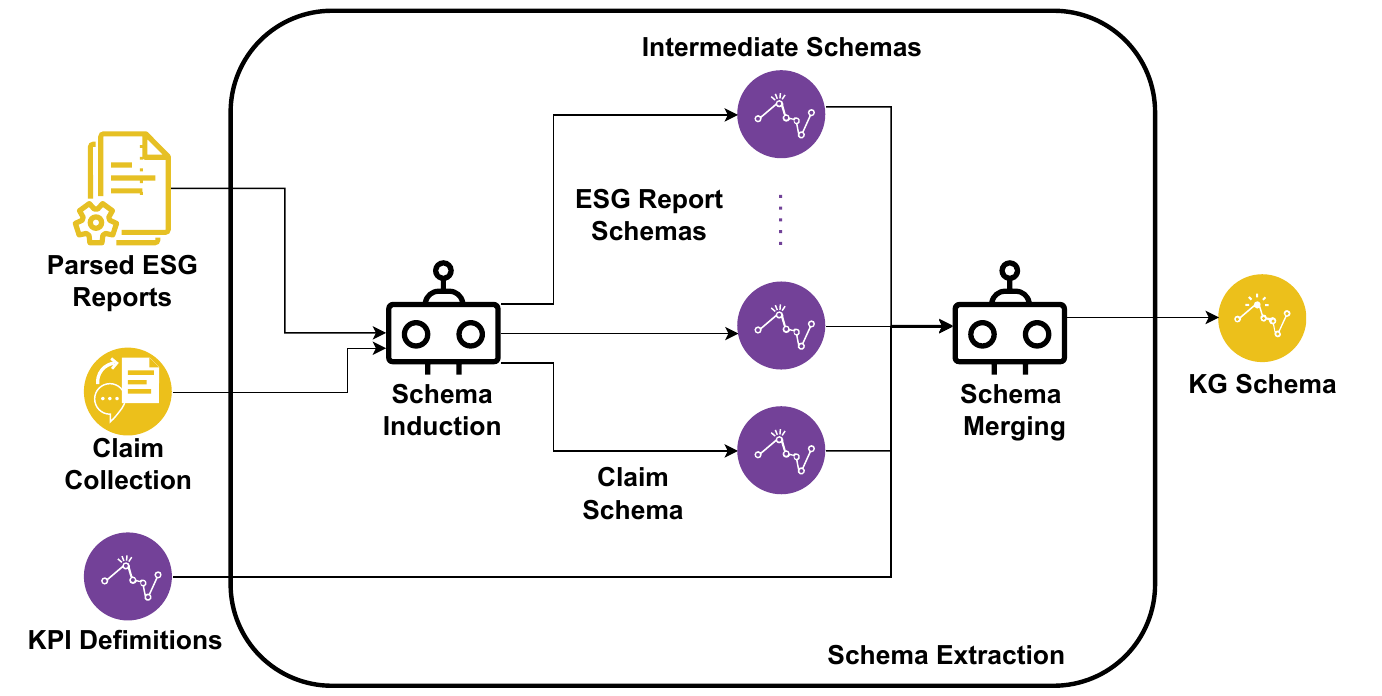}
    \caption{Schema Extraction}
    \label{fig:schema}
\end{figure}

The schema $S$ is synthesized from three sources as depicted in Figure~\ref{fig:schema}: (i) data-driven patterns observed in the parsed ESG corpus, yielding a merged schema $S_{\rm data}$ of frequent entity–relation configurations induced from each report, (ii) a regulatory sub-schema $S_{\rm reg}$ of key performance indicators derived from official standards, and (iii) a claim-driven schema $S_{\rm claim}$ constructed from known greenwashing examples by abstracting common claim patterns. These partial schemas are merged into a unified schema $S = S_{\rm data} \cup S_{\rm reg} \cup S_{\rm claim}$ that specifies allowed entity types (e.g., \texttt{Company}, \texttt{Facility}, \texttt{KPIObservation}, \texttt{SustainabilityClaim}) and relation types (e.g., \texttt{reportsKPI}, \texttt{setsGoal}) along with their attribute domains. One key design choice is a company-centered schema, where a single \texttt{Organization} node anchors all entities and relations for each company, normalizing heterogeneous disclosures around a consistent corporate identity. For example, all emission reports by Company X attach to that \texttt{Organization} node, ensuring structural clarity. This schema-based modeling enables efficient context retrieval, constrains inference to valid schema patterns for company-centered claims, and supports explainable evidence trails.

\begin{figure}[ht!]
    \centering
    \includegraphics[width=0.9\linewidth]{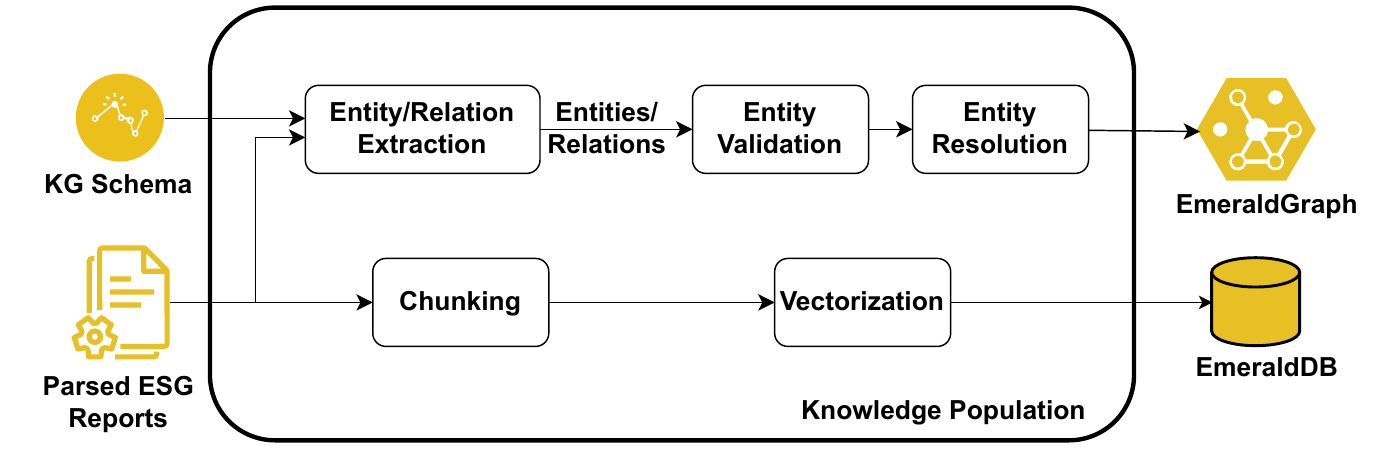}
    \caption{Knowledge Population}
    \label{fig:knowledgePopulation}
\end{figure}

\paragraph{Knowledge population.}
Using the parsed reports and extracted schema, we  instantiate the \emph{EmeraldDB} and \emph{EmeraldGraph} stores with actual data from reports as depicted in Figure~\ref{fig:knowledgePopulation}.
To populate the \emph{EmeraldDB}, each report is segmented into chunks $d_i$, and an embedding is computed for each. We store for each chunk: $d_i$=$(\text{embedding}, \text{reportID}, \text{company},$ $\text{year}, \text{chunkID}, \text{page\_number})$. Chunks have a 250-token size, with a 50-token overlap to preserve contextual meaning, while enabling effective reasoning.

\begin{figure}[ht!]
    \centering
    \includegraphics[width=0.9\linewidth]{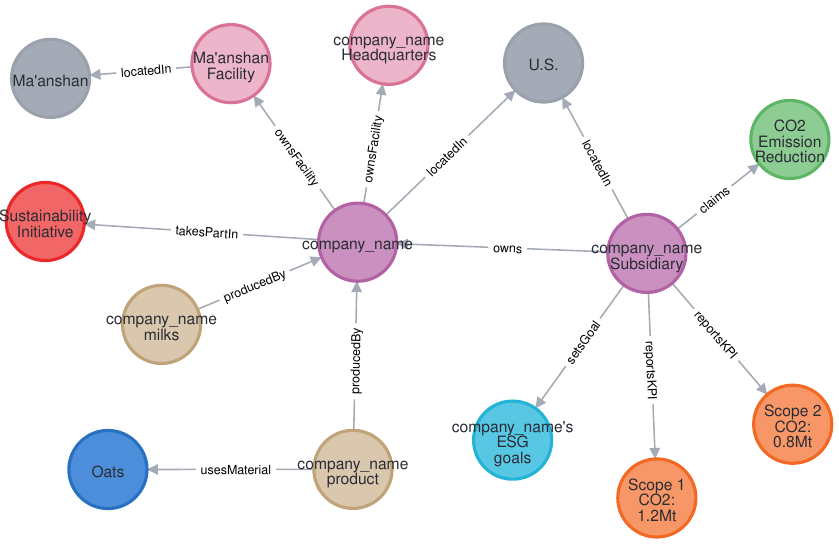}
    \caption{EmeraldGraph snippet centered on an anonymized company. Node types in the EmeraldGraph knowledge graph are color-coded as follows: \textcolor[HTML]{F16667}{Initiative}, \textcolor[HTML]{A5ABB6}{Location}, \textcolor[HTML]{4C8EDA}{Material}, \textcolor[HTML]{D9C8AE}{Product}, \textcolor[HTML]{23b3d7}{Goal}, \textcolor[HTML]{8DCC93}{Sustainability Claim}, \textcolor[HTML]{C990C0}{Organization}, \textcolor[HTML]{ECB5C9}{Facility}, \textcolor[HTML]{F79767}{KPIObservation}.}
    \label{fig:emeraldgraph_anon}
\end{figure}

To populate the \emph{EmeraldGraph}, the report content is mapped to graph entities and relations with respect to the schema $S$. 
Challenges here include extraction accuracy and ensuring that each real-world entity is represented by a unique node  (“ABC Corp.” vs.\ “ABC Corporation”). We mitigate this by mapping noisy document mentions to schema-typed entities and relations and by using embedding-blocking \cite{li2024entity,wang2024comem,jiang2024chatea,peeters2024entity} to avoid duplicate or misaligned segments.
For each parsed ESG report, an LLM extracts candidate triples $(u,\ell,v)$. 
A candidate triple $(u,\ell,v)$ is only admitted if its type is valid against the schema $S$ under $\tau(u) \xrightarrow{\ell} \tau(v)$. 
Figure~\ref{fig:emeraldgraph_anon} shows a real-world snippet from EmeraldGraph centered on an anonymized company.

\section{Knowledge-powered Reasoning}
\label{sec:inference}
\begin{figure}[ht]
    \centering
    \includegraphics[width=0.9\linewidth]{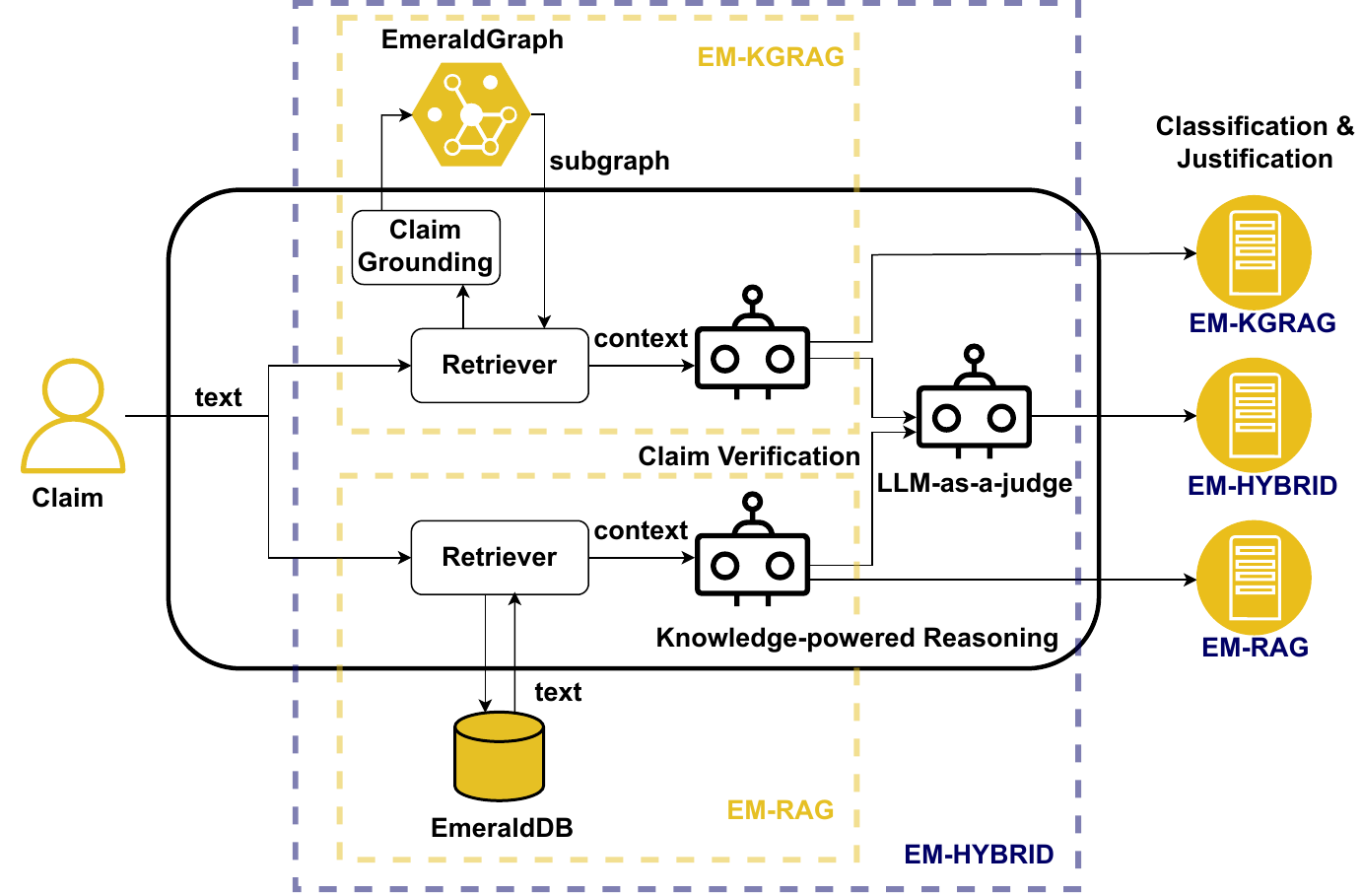}
    \caption{Knowledge-powered Reasoning}
    \label{fig:inference}
\end{figure}

In this section, we present the Knowledge-powered Reasoning phase (Figure~\ref{fig:inference}), where a textual sustainability claim $c$ (e.g., “\emph{Company X reduced its CO$_2$ emissions by 30\% in 2023}”) is evaluated using evidence from \emph{EmeraldGraph} and \emph{EmeraldDB}. We consider three pipeline configurations:  \texttt{EM-KGRAG}, which uses only graph evidence $H$ from \emph{EmeraldGraph}; \texttt{EM-RAG}, which uses only document chunks $\{d_i\}$ from \emph{EmeraldDB}; and \texttt{EM-HYBRID}, which uses as context the justifications produced by \texttt{EM-KGRAG} and \texttt{EM-RAG}.

\paragraph{Claim Grounding.}
Given a claim $c$, Claim Grounding identifies the target company and parses $c$ for other key elements: KPIs, numeric values, policy mentions, and goals. 
Then, it links the identified company to $v_{company}$, and the other identified key elements to specific nodes and types in the \emph{EmeraldGraph}.
Each extracted element is mapped to a schema type in $S$ (e.g. \texttt{KPIObservation}, \texttt{Policy}, \texttt{Goal}) and the node's attributes are populated using the property function $p$. For example, in the claim “\emph{Company X reduced its CO$_2$ emissions by 30\% in 2023}”, we would link “Company X” to the $v_{company}$, and resolve the “30\%” emission into a new node $v_{observation}$ connected to $v_{company}$, where $\tau(v_{obs}) = \texttt{KPIObservation}$, and $p(v_{observation}) = \{ \text{value}: 30, \text{unit}: \%, \text{year}: 2023 \}$.
Claim grounding outputs the resolved $v_{company}$ node and a node set $V_{claim}$ of the key-elements defined by their specific types $\tau$ and property maps $p$.

We identify two key challenges in claim grounding. First is its linguistic variability and ambiguity. Sustainability claims often use imprecise language (e.g., “net-zero” vs.\ a concrete target, or “our emissions” without a unit or year) and industry-specific terminology. For example, the term “flaring” must be recognized as “burning gas” and thus “CO$_2$ emissions”; otherwise, the system might store it as a \textsc{SafetyIncident} instead of an \textsc{EmissionObservation}. Mapping such phrases to precise schema types requires robust LLM-based parsing. We therefore employ a schema-based retrieval algorithm that restricts retrieval to the types explicitly referenced in the claim, focusing reasoning on concrete entities and avoiding unrelated graph regions. Second, the claims do not necessarily include  KPI elements, which are imperative for the retrieval of quantifiable evidence. To mitigate this, we add a node $u$ to $V_{claim}$ with $\tau(u)=$\textsc{KPIObservation} using the claim as attribute, ensuring that we always query \textsc{KPIObservation} nodes. 

\paragraph{Schema-Based Context Retrieval.}
\label{sec:retrieval}
Schema-Based Context Retrieval performs schema-based subgraph extraction from \emph{EmeraldGraph} to build a graph evidence context. With the claim grounded in graph nodes, we first retrieve a subgraph containing relevant evidence to form the context for classification.
Using the grounded company node $v_{\text{company}}$ and the set of schema types $\{\tau_j\}$ identified in $c$, we construct an evidence subgraph $H \subseteq G$. For each type $\tau_j$, Algorithm~\ref{alg:evidence_retrieval} performs a breadth-first search from $v_{\text{company}}$ up to $k$ hops, collecting all nodes of type $\tau_j$. It then ranks these candidates by cosine similarity between their node embeddings and the corresponding key elements found in the claim, retains those above a threshold, and keeps the top-$n$ most similar nodes per type. For each selected node $v$, we compute the shortest path $P(v_{\text{company}}, v)$ and add all nodes and edges on these paths to $H$. The resulting context subgraph consists of reasoning paths rather than isolated facts.

A key challenge is to extract a compact context, i.e., a claim-specific evidence subgraph from \emph{EmeraldGraph}. Naïve neighborhood expansion around the $v_{\text{company}}$ either floods the context with loosely related nodes or omits multi-hop facts that matter for greenwashing (e.g., links to nodes topologically deeper in the graph, such as \textsc{Material}). Schema-Based Context Retrieval addresses this by constraining traversal to the entity types explicitly referenced in the claim and by limiting search to $k$-hop neighborhoods around the grounded $v_{\text{company}}$. Candidate nodes are then filtered by embedding-based similarity to the claim anchors, so only semantically relevant instances are kept. Finally, we add shortest paths between the company node and selected nodes, producing concise reasoning chains rather than arbitrary neighborhoods.

\paragraph{Document Retrieval.}
Document Retrieval retrieves textual evidence from \emph{EmeraldDB}. Using the company name extracted from the claim, we use the identifier $company$ to limit our search to relevant chunks in the document store. We compute the embedding of the claim $c$ and measure cosine similarity to each chunk embedding in the store. We then return the top-$m$. 
Restricting to the target company avoids irrelevant documents and ensures that retrieved chunks are contextually aligned with the claim, improving factual grounding.

\paragraph{Classification and Justification.}
In \texttt{EM-KGRAG} and \texttt{EM-RAG} variants, the claim $c$ and its corresponding context are fed into a prompted LLM, which
performs contextual reasoning to produce the final verdict (\textsc{Greenwashing}, \textsc{Not Greenwashing}, \textsc{Abstain}) and a factual justification for the classification label.
The \texttt{EM-HYBRID} variant requires the justification outputs from \texttt{EM-KGRAG} and \texttt{EM-RAG}, which serve as context for LLM reasoning; here, the verifier acts as a judge that selects one of the two justifications and outputs it along with the corresponding label.

\begin{algorithm}[t]
\caption{Schema-driven Context Retrieval}
\small
\label{alg:evidence_retrieval}
\begin{algorithmic}[1]
\Require $v_{company}$ (organization node ID), 
$\mathcal{V}_{claim}$ (nodes extracted from claim grounding),
$top\_n$ (no. retrieved nodes per type), 
$threshold$ (similarity threshold), 
$k$ (max path length in hops)
\Ensure $H$ (subgraph with retrieved nodes and edges) 
    \State $H \leftarrow \emptyset$ 
    \ForAll{$v \in V_{claim}$}
       \State $V_r \leftarrow \textsc{Retrieve}( v_{company}, \tau(v), k)$ \Comment{Collect all nodes of the specified type within company's k-hop neighborhood}
        \State $V_{sim} \leftarrow \emptyset$ \Comment{Nodes that pass the similarity threshold}
        \ForAll{$u$ in $V_r$} \Comment{Compute similarities and select top-n nodes}
            \State $sim \leftarrow \textsc{CosSim}(\textsc{Embedding}(v), \textsc{Embedding}(u))$
            \If{$sim \geq threshold$}
                \State $V_{sim} \leftarrow V_{sim} \cup \{(u, sim)\}$
            \EndIf
        \EndFor
        \State $V_{tn} \leftarrow \textsc{TopN}(\textsc{Sort}(V_{sim}), top\_n)$ \Comment{Top-n most similar nodes}
        \ForAll{$u \in V_{tn}$} \Comment{Compute the shortest paths for top-n nodes}
            \State $path \leftarrow \textsc{ShortestPath}(v_{company},u$)
            \State $H \leftarrow H \cup \{ path \}$
        \EndFor
    \EndFor
    \State \Return $H$
\end{algorithmic}
\end{algorithm}

\section{The EmeraldData Benchmark}
\label{sec:datasets}

A fundamental limitation of existing research on greenwashing is the absence of large-scale annotated real-world benchmarks containing verified instances of greenwashing \cite{calamai2025corporate}.
Several factors contribute to this scarcity, including the vagueness of greenwashing definitions, context-sensitive claims, annotation complexity requiring domain expertise, and legal and reputational implications of labeling corporate claims as deceptive. 

To the best of our knowledge, the only relevant publicly available greenwashing detection dataset is  \texttt{GreenClaims} \cite{greenwashingdataset}. It contains only a limited number of 91 claim samples, from which only 51 were usable in our evaluation due to the availability of corresponding ESG reports. To overcome this limitation, we introduce \texttt{EmeraldData}, a larger semi-synthetic dataset (620 instances), constructed via a four-stage pipeline inspired by \cite{wright2022generating,pan2021zero,bussotti2024unknown}. 
First, we used the smaller \texttt{GreenClaims} benchmark to extract 37 (company, year) unique pairs. These allow us to align claims with the ESG reports used for the creation of \texttt{EmeraldMind} Evidence Stores.  
Second, we collect relevant articles from reliable news sites with topics including, but not limited to, greenwashing, sustainability news, ESG news, company news regarding various ESG goals, or any accusations and litigations that a company may face. We filter them by (company, year) pairs as defined in the first step, ensuring contextual relevance. Third, we prompt an LLM with article metadata to generate both truthful (non-greenwashing) and refuting (greenwashing) claims, yielding 620 candidate instances. Finally, the same model assigns a label to each claim and produces a brief textual justification anchored to the source article, enabling transparent, article-grounded evaluation.
Table~\ref{tab:datasets} summarizes the datasets. 
\begin{table}[ht!]
    \centering
      \caption{Summary of the datasets used in the experimental evaluation with G as the number of Greenwashing and NG as the number of Not Greenwashing.}
    \label{tab:datasets}
    \begin{tabular}{lccc}\hline
            \textbf{Dataset}& \textbf{No. Claims} & \textbf{G} & \textbf{NG} \\ \hline
        \texttt{GreenClaims} & 51 & 24 (47\%)& 27 (53\%)\\ 
        \texttt{EmeraldData}  & 620 & 225 (36\%)& 395 (64\%) \\ \hline
            \end{tabular}
 \end{table}

\section{Experimental Evaluation }
\label{sec:exp}

\subsection{Experimental Setup} 
We evaluated all \texttt{EmeraldMind}  variants: \texttt{EM-RAG} ; \texttt{EM-KGRAG} (statistics in Section~\ref{sec:graphStats}); and \texttt{EM-HYBRID}, benchmarked against a baseline LLM relying solely on internal knowledge under two prompt configurations. The first (zero-shot prompt) uses only claim text and greenwashing definitions from the EU Green Claims Directive \cite{EuropeanCommission2023GreenClaimsDirective}. The second (few-shot prompt) adds few-shot examples found in \cite{calamai2025corporate}. 
This comparison demonstrates how domain-specific evidence improves trustworthiness over generic learned patterns, aligning with responsible AI principles. Evaluation metrics include classification accuracy, coverage, and justification quality (Section~\ref{sec:results}). 

The retrieval hyperparameters are in  Table~\ref{tab:hyper}. Experiments were conducted on a system equipped with an NVIDIA RTX A6000 GPU with dual AMD EPYC 9335 32-core processors and 128 GB RAM, using \texttt{gemma-27b-it} for inference, \texttt{prometheus-13b-v1.0} for ILORA evaluation, and for ranking/hybrid judging \texttt{prometheus-7b-v2.0}. Prometheus \cite{kim2023prometheus} is an open-source evaluator LLM designed for reproducible, fine-grained assessment as a practical alternative to human evaluation.
{\textbf{Retrieval Hyperparameters}}
The default hyperparameters are summarized in Table~\ref{tab:hyper}.
\begin{table}[H]
\centering
\caption{Default Hyperparameters for Retrieval}
\label{tab:hyper}
\begin{small}
    \begin{tabular}{llcc}
\hline
\textbf{Pipeline} & \textbf{Symbol} &\textbf{Hyperparameter} & \textbf{Value} \\
\hline
\multirow{3}{*}{\texttt{EM-KGRAG}} & $top_n$ & No. nodes retrieved per type & 3  \\
& $\tau$ & Similarity Threshold & 0.2  \\
 & $k$ & Max path length (hops)  & 3\\ \hline
\multirow{1}{*}{\texttt{EM-RAG}} & $top_m$ & No. chunks retrieved & 8 \\
\hline
\end{tabular}
\end{small}

\end{table}

\subsection{EmeraldGraph Knowledge Graph}
\label{sec:graphStats}
We constructed \textit{EmeraldGraph} using the pipeline described in Section~\ref{sec:stores}, extracting structured facts from 37 publicly available ESG reports. 
Table~\ref{tab:graph_stats} shows that \emph{EmeraldGraph} is a sparse, company-centered graph. The 53{,}748 entities and 59{,}344 relationships indicate low edge density and predominantly small local neighborhoods. The centrality statistics by node type reveal that \textsc{Organization} nodes act as high-degree hubs in a core--periphery topology. This induces star-like networks centered on organizations, with peripheral nodes such as facilities and locations attaching to a small number of central company nodes. The very large average shortest-path length and diameter further suggest that global connectivity is weak and that effective reasoning should be localized around these organizational cores, as long paths or disconnected regions make global traversal inefficient.

\begin{figure*}[ht!]
        \centering
       \includegraphics[width=0.9\textwidth]{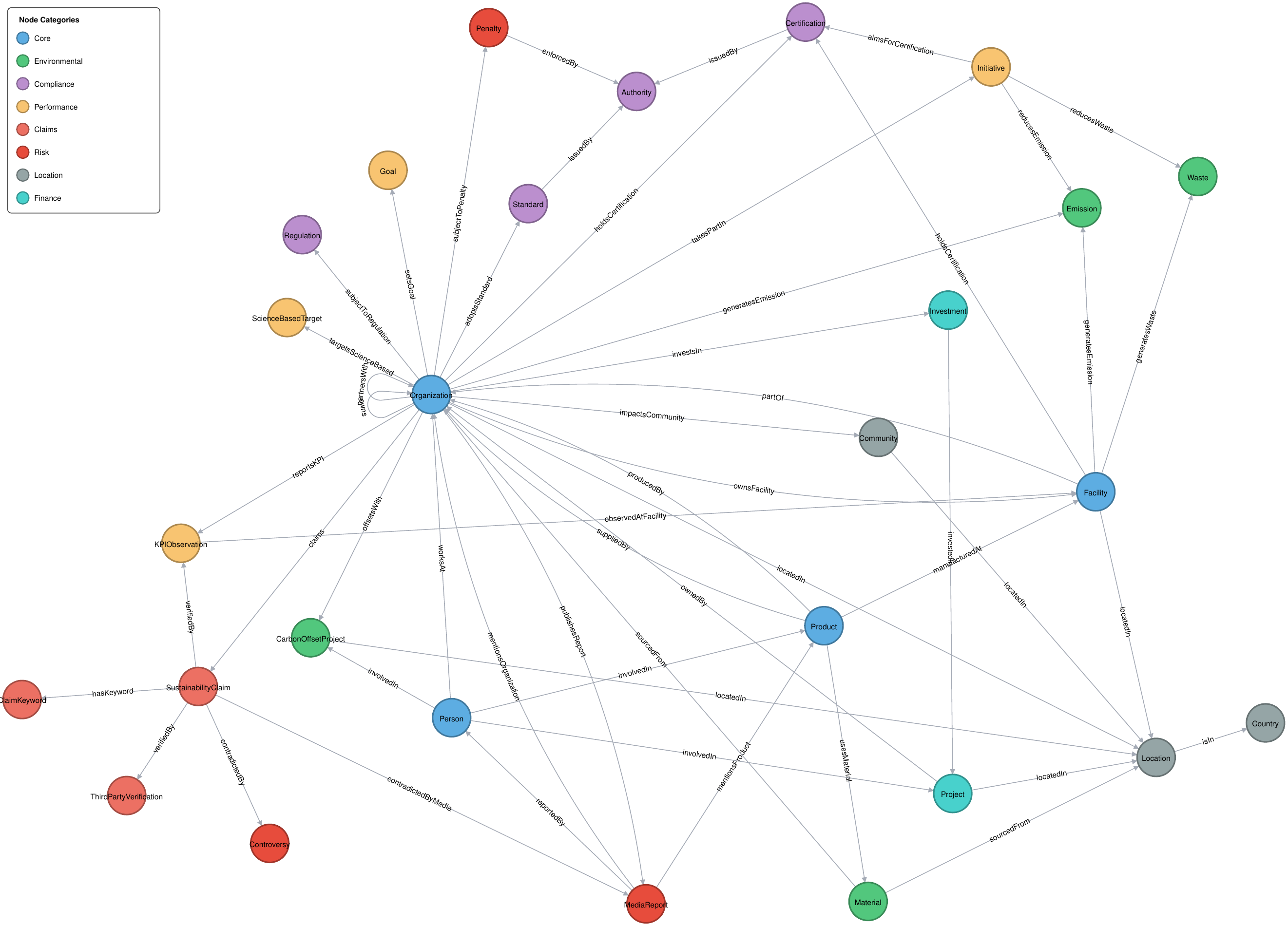}
        \caption{EmeraldGraph Schema}
        \label{fig:esg-schema}
\end{figure*}

Table~\ref{tab:top_distributions} highlights how the graph density is concentrated on KPI- and claim-related structure. \textsc{KPIObservation} accounts for 24{,}809 nodes, which is approximately 46\% of all entities, and \texttt{reportsKPI} for 24{,}832 edges, which is approximately 42\% of all relationships. This implies dense connectivity between the two entity types. The remaining frequent entity types define the dominant local motifs around each company, capturing performance reporting, goal setting, initiatives, and geographic scope. Overall, the degree distribution and type frequencies confirm that \emph{EmeraldGraph} allocates most of its structural capacity to modeling company-centric KPI observations, claims, and goals, which are precisely the regions exploited by our schema-based retrieval algorithm. The entire \emph{EmeraldGraph} schema is available in Figure~\ref{fig:esg-schema}.

\begin{table*}[ht!]
    \centering
    \caption{EmeraldGraph Statistics}
    \label{tab:graph_stats}
    \begin{small}
    \begin{tabular}{lrclr}
         \cline{1-2} \cline{4-5}
        \multicolumn{2}{c}{\textbf{Graph Metrics}} && \multicolumn{2}{c}{\textbf{Centrality by Node Type}} \\
        \cline{1-2} \cline{4-5}
        \textbf{Metric} & \textbf{Value} && \textbf{Node Type} & \textbf{Avg.\ Degree} \\
        \cline{1-2} \cline{4-5}
        No. of Entities & 53,748 && Organization & 17.95 \\
        No. of Relationships & 59,344 && Country & 6.40 \\
        Avg.\ Total Degree & 2.21 && Location & 3.43 \\
        Avg.\ Shortest Path Length & 1,727.83 && Facility & 2.63 \\
        Diameter & 23,788 && Material & 1.77 \\
         \cline{1-2} \cline{4-5}
    \end{tabular}
    \end{small}
\end{table*}

\begin{table}[ht!]
    \centering
    \caption{Top-5 Entity and Relationship Types by Frequency}
    \label{tab:top_distributions}
    \begin{small}
    \begin{tabular}{lrclr}
         \cline{1-2} \cline{4-5}
        \multicolumn{2}{c}{\textbf{Top-5 Entities}} && \multicolumn{2}{c}{\textbf{Top-5 Relationships}} \\
        \cline{1-2} \cline{4-5}
        \textbf{Entity Type} & \textbf{Count} && \textbf{Relation Type} & \textbf{Count} \\
          \cline{1-2} \cline{4-5}
        KPIObservation & 24,809 &&reportsKPI & 24,832 \\
        Initiative & 4,060 && takesPartIn & 4,388 \\
        SustainabilityClaim & 3,458 && setsGoal & 3,475 \\
        Goal & 3,414 && claims & 3,446 \\
        Organization & 3,020 && locatedIn & 3,396 \\
         \cline{1-2} \cline{4-5}
    \end{tabular}
    \end{small}
\end{table}

\subsection{Experimental Results}
\label{sec:results}

\subsubsection*{\textbf{Classification Performance}}
To evaluate \texttt{EmeraldMind}, we report accuracy, coverage, overall accuracy (accuracy$\times$coverage), and abstentions against ground-truth annotations, providing insight into predictive performance and selective behavior. Table~\ref{tab:stats}
summarizes these metrics over the two datasets. 
\begin{table*}[ht]
    \centering
    \caption{Classification performance across benchmarks, pipelines, and prompt variations. Accuracy and Coverage measure conditional performance on non-abstained claims; Overall Acc. reflects accuracy including abstentions ($= \text{Accuracy} \times \text{Coverage}$).}
    \label{tab:stats}
   \begin{tabular}{lllcccc}
    \hline
    \textbf{Dataset} & \textbf{Prompt} & \textbf{Pipeline} &
     \textbf{Accuracy} & \textbf{Coverage} & \textbf{Overall Acc.} & \textbf{No. Abstains} \\
    \hline

    \multirow{8}{*}{\texttt{GreenClaims}}
    & \multirow{4}{*}{Zero-shot}
        & \texttt{Baseline}   & 93.33\% & 29.41\% & 27.45\% & 36 \\
    &   & \texttt{EM-RAG}     & 82.14\% & 54.90\% & 45.10\% & 23 \\
    &   & \texttt{EM-KGRAG}   & 93.55\% & 60.78\% & 56.86\% & 20 \\
    &   & \texttt{EM-HYBRID}  & 89.47\% & 74.51\% & 66.67\% & 13 \\ \cline{2-7}
    & \multirow{4}{*}{Few-shot}
        & \texttt{Baseline}   & 100.00\% & 31.37\% & 31.37\% & 35 \\
    &   & \texttt{EM-RAG}     & 77.14\%  & 76.47\% & 52.94\% & 12 \\
    &   & \texttt{EM-KGRAG}   & 89.74\%  & 76.47\% & 68.63\% & 12 \\
    &   & \texttt{EM-HYBRID}  & 92.31\%  & 76.47\% & 70.59\% & 12 \\
    \hline

    \multirow{8}{*}{\texttt{EmeraldData}}
    & \multirow{4}{*}{Zero-shot}
        & \texttt{Baseline}   & 94.21\% & 25.97\% & 24.52\% & 459 \\
    &   & \texttt{EM-RAG}     & 87.92\% & 62.74\% & 55.16\% & 231 \\
    &   & \texttt{EM-KGRAG}   & 92.51\% & 49.52\% & 45.81\% & 313 \\
    &   & \texttt{EM-HYBRID}  & 85.78\% & 68.06\% & 58.39\% & 198 \\ \cline{2-7}
    & \multirow{4}{*}{Few-shot}
        & \texttt{Baseline}   & 94.21\% & 19.52\% & 18.39\% & 499 \\
    &   & \texttt{EM-RAG}     & 85.19\% & 69.68\% & 59.35\% & 188 \\
    &   & \texttt{EM-KGRAG}   & 88.03\% & 60.65\% & 53.39\% & 244 \\
    &   & \texttt{EM-HYBRID}  & 83.80\% & 74.68\% & 62.58\% & 157 \\
    \hline
\end{tabular}
\end{table*}

\texttt{EmeraldMind} variants substantially outperform the baseline in decision coverage across both datasets, achieving 2-4 times higher rates (49-77\% vs. 19-31\%) while maintaining competitive accuracy (77-93\% vs. 93-100\%). Overall accuracy, accounting for abstentions, shows that \texttt{EM-HYBRID} achieves the highest effective performance (up to 70.59\% on \texttt{GreenClaims} few-shot), as it leverages LLM-as-a-judge pairwise comparison to select superior justifications from \texttt{EM-RAG} and \texttt{EM-KGRAG}, suggesting that a good quality justification leads to more accurate classification.
Few-shot prompting consistently boosts the coverage of the  \texttt{EmeraldMind} pipelines compared to zero-shot, though baseline coverage remains stagnant or declines. This confirms RAG-based reasoning benefits from example-guided reasoning, reducing abstentions.
On the smaller \texttt{GreenClaims} dataset (51 claims), \texttt{EM-KGRAG} excels with the highest zero-shot accuracy (93.55\%) and few-shot overall accuracy (68.63\%). Conversely, \texttt{EM-RAG} dominates the larger \texttt{EmeraldData} (620 claims) with superior coverage (62-77\%) and overall accuracy (55-62\%). \texttt{EM-HYBRID} maximizes performance in both by successfully combining the two variants' performance strengths.

\subsubsection*{\textbf{Justification Quality}}

The metrics in Table~\ref{tab:stats} do not fully capture the quality of reasoning behind the justification generation. Prior research has shown that classification metrics do not assess hallucination, factual consistency, or the soundness of model reasoning \cite{dang2025survey}. 
To address this, we follow approaches that incorporate LLM-driven qualitative assessment \cite{gu2024survey}.
We adopt an LLM-as-a-judge paradigm \cite{zheng2023judging} with two complementary evaluation strategies. 
First, \textit{single answer grading} (\textit{absolute LLM judge} \cite{sahoo2025quantitative}), which scores each \texttt{EmeraldMind} variant and baseline justification independently based on established criteria. Second, \textit{pairwise comparison} (\textit{relative LLM judge} \cite{sahoo2025quantitative}), which compares justifications from \texttt{EM-RAG}, \texttt{EM-KGRAG}, and the baseline directly. Since \texttt{EM-HYBRID} selects the higher-ranked justification between \texttt{EM-RAG} and \texttt{EM-KGRAG} from pairwise comparisons, we exclude it from this evaluation.

\begin{figure}[ht]
    \centering
    \begin{subfigure}[b]{0.4\textwidth}
        \includegraphics[width=\textwidth]{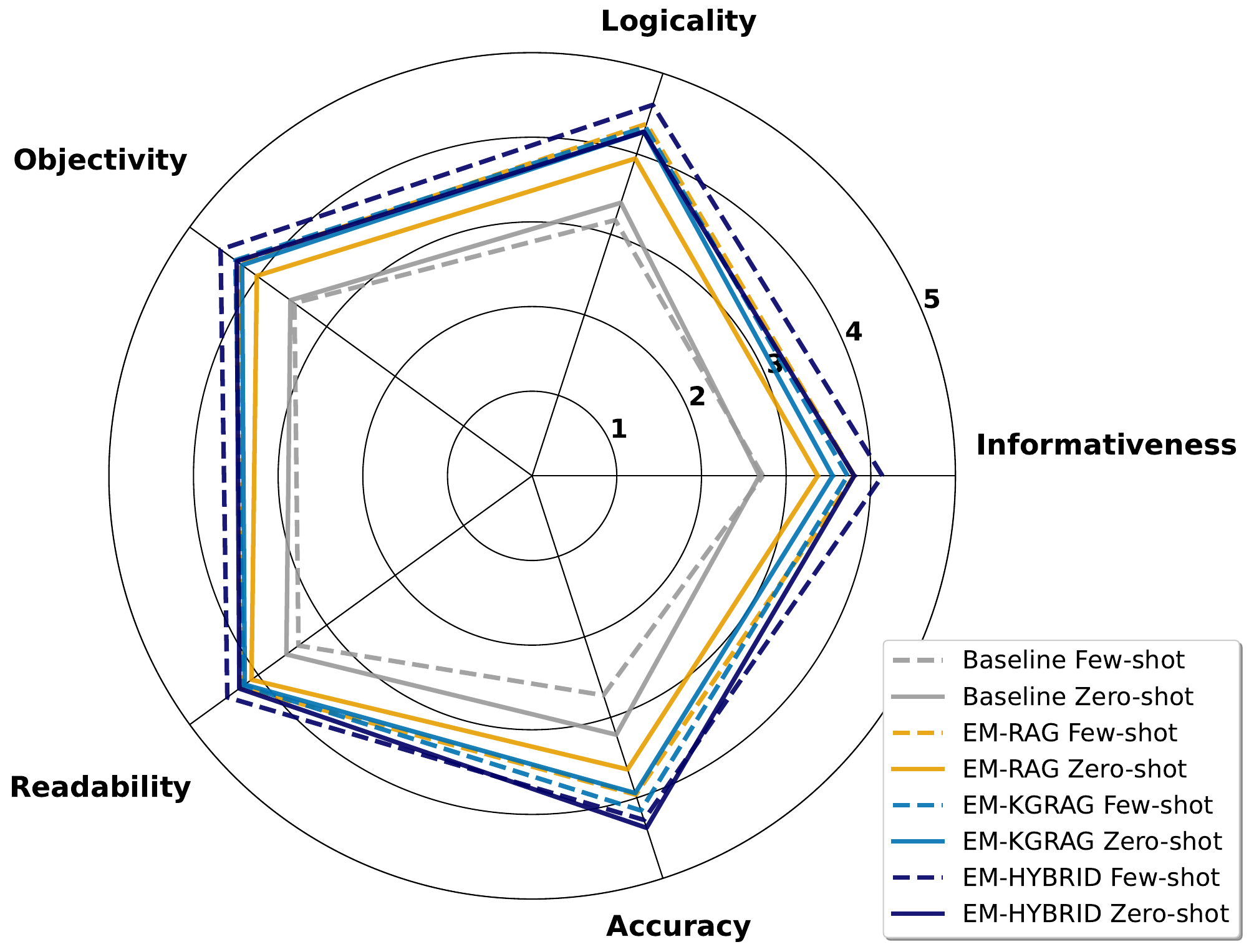}
        \caption{\texttt{GreenClaim}}
    \end{subfigure}
    \begin{subfigure}[b]{0.4\textwidth}
        \includegraphics[width=\textwidth]{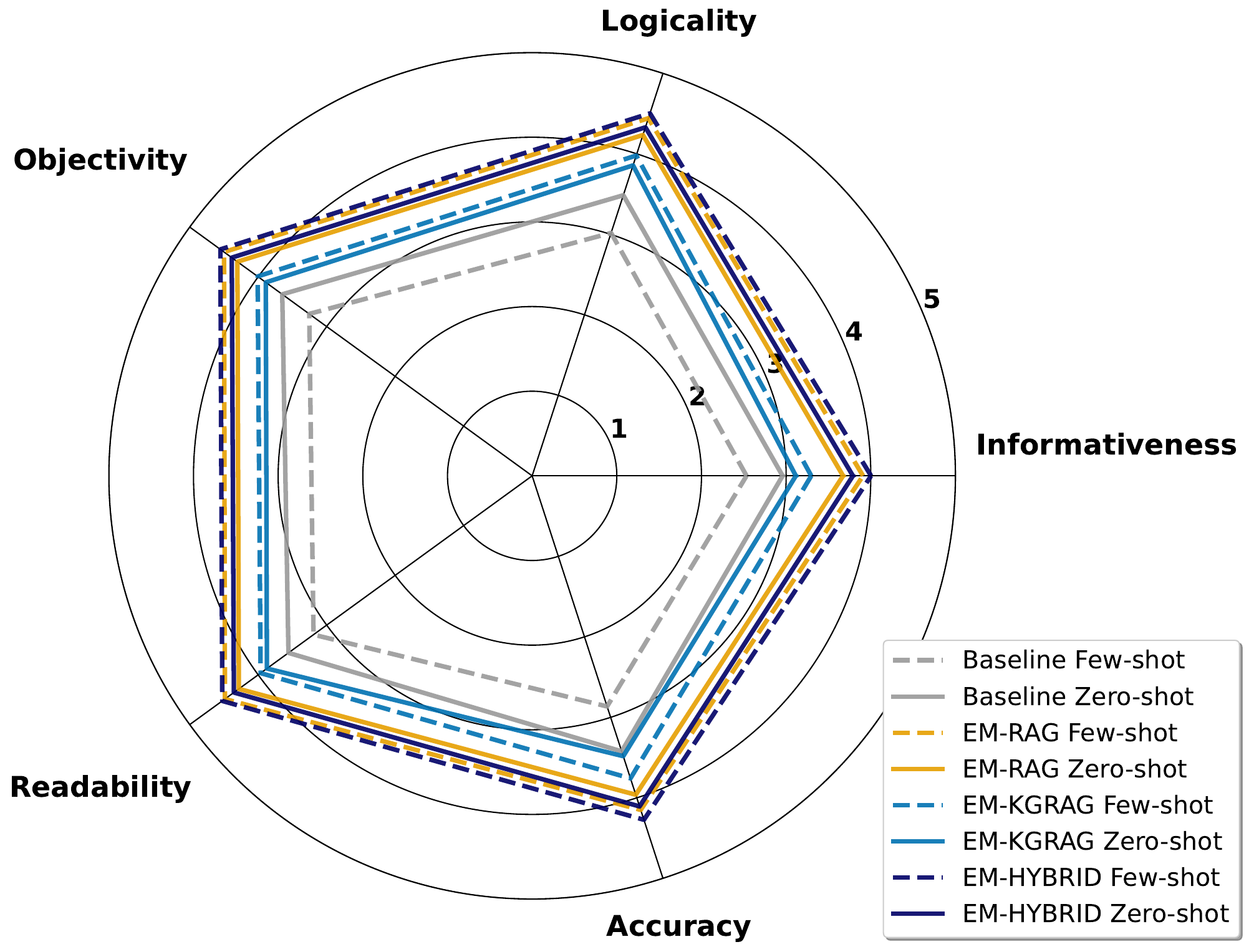}
        \caption{\texttt{EmeraldData}}
    \end{subfigure}
    \caption{ILORA scores for the different prompt-pipeline combinations, highlighting performance on key metrics. Each metric is rated on a 5-point scale, where 1 indicates the lowest quality and 5 the highest.
}
    \label{fig:ilora}
\end{figure}

\noindent{\textbf{(1) Single Answer Grading.}}
For the evaluation of the justifications, we use ILORA \cite{zheng2025predictions},  an Explanation Quality Evaluation Method employing a 5-point Likert scale across five criteria:  \textbf{Informativeness (I)} - provision of new information, such as background or additional
context; \textbf{Logicality (L)} - coherent reasoning process and causal link to the outcome;  \textbf{Objectivity (O)} - objectivity of the answer and bias-free analysis; \textbf{Readability (R)} - grammatical correctness, structural clarity, and ease of comprehension; \textbf{Accuracy (A)} - alignment with the true label and its accurate reflection of the result.
Figure~\ref{fig:ilora} shows radar charts comparing ILORA scores across prompt-pipeline combinations.

In both datasets, \texttt{EM-RAG} and \texttt{EM-KGRAG} consistently outperform the baseline across all ILORA metrics. Few-shot prompting improves explanation quality over zero-shot across all \texttt{EmeraldMind} variations, while the baseline's limited knowledge hampers its ability to provide factual explanations, leading to more abstentions and lower quality. \texttt{EM-HYBRID} achieves the highest ILORA scores across all criteria and datasets, under corresponding prompting conditions, demonstrating coherence between individual scoring and pairwise comparisons, as it selects the superior justification between \texttt{EM-RAG} and \texttt{EM-KGRAG}.

Specifically, in the \texttt{GreenClaim} benchmark (average scores depicted in Figure~\ref{fig:ilora}(a)), \texttt{EM-KGRAG} excels in Objectivity and Accuracy under the few-shot setting. Conversely, in the larger \texttt{EmeraldData} benchmark (Figure~\ref{fig:ilora}(b)), \texttt{EM-RAG} outperforms \texttt{EM-KGRAG} across all ILORA metrics, showcasing superior explanation quality under both prompting conditions. The baseline consistently exhibits the lowest performance, especially in Readability and Logicality. These results confirm that few-shot prompting enhances reasoning depth and explanation reliability across datasets and models.

\noindent\textbf{(2) Pairwise Comparison.}
We conduct relative evaluation using a 3-way LLM-as-a-Judge instead of pairwise. Specifically, an LLM ranks all three justifications (\texttt{Baseline}, \texttt{EM-RAG}, \texttt{EM-KGRAG}) simultaneously under ILORA metrics. Table~\ref{tab:rankings} presents the count of first, second, and third place rankings for each pipeline across dataset-prompt combinations using the relative judge paradigm outputs. To derive an overall ranking, the Borda count \cite{fishburn1976borda} method is applied, assigning 3 points for first place, 2 points for second, and 1 point for third place. For all the dataset-prompt variation, the resulting Borda scores (see Table~\ref{tab:borda}) consistently yield the ranking: 
\begin{center}
\texttt{EM-RAG} $\succ$ \texttt{EM-KGRAG} $\succ$ \texttt{Baseline}.
\end{center}

Table~\ref{tab:borda} presents the Borda scores derived from the \emph{relative LLM judge} for all dataset-prompt variations.
\begin{table}[H]
\centering
\caption{Borda Scores}
\label{tab:borda}
\begin{small}
\begin{tabular}{llccc}
\hline
\textbf{Dataset} & \textbf{Prompt} & \textbf{\texttt{Baseline}} & \textbf{\texttt{EM-RAG}} & \textbf{\texttt{EM-KGRAG}} \\
\hline
\multirow{2}{*}{\texttt{Greenclaims}} 
 & Zero-shot & 64 & 140 & 102 \\ 
 & Few-shot & 52 & 141 & 113 \\ \hline
\multirow{2}{*}{\texttt{EmeraldData}} 
 & Zero-shot & 875 & 1788 & 1057 \\ 
 & Few-shot & 667 & 1787 & 1266\\ \hline
& Summary& 1658 & 3856 & 2538\\ \hline
\end{tabular}
\end{small}
\end{table}

\begin{table}[t!]
\centering
\caption{Comparison of Baseline, RAG, and GraphRAG counts grouped by dataset-prompt.}
\label{tab:rankings}
\begin{small}
\begin{tabular}{lllccc}
\hline
\textbf{Dataset} &\textbf{Prompt} & \textbf{Pipeline} & \textbf{1st} & \textbf{2nd} & \textbf{3rd} \\
\hline
\multirow{6}{*}{\texttt{Greenclaims}} 
 &\multirow{3}{*}{Zero-shot} &  \texttt{Baseline}  & 1 & 11 & 39 \\
 && \texttt{EM-RAG} & 38 & 13 & 0 \\
 && \texttt{EM-KGRAG} & 12 & 27 & 12 \\
\cline{2-6}
&\multirow{3}{*}{Few-shot} &  \texttt{Baseline}  & 0 & 1 & 50 \\
 && \texttt{EM-RAG} & 39 & 12 & 0 \\
 && \texttt{EM-KGRAG} & 12 & 38 & 1 \\
\hline
\multirow{6}{*}{\texttt{EmeraldData}} 
 &\multirow{3}{*}{Zero-shot} &  \texttt{Baseline}  & 12 & 231 & 377 \\
 && \texttt{EM-RAG} & 553 & 62 & 5 \\
 && \texttt{EM-KGRAG} & 55 & 327 & 238 \\
\cline{2-6}
&\multirow{3}{*}{Few-shot} &  \texttt{Baseline}  & 13 & 21 & 586 \\
 && \texttt{EM-RAG} & 551 & 65 & 4 \\
 && \texttt{EM-KGRAG} & 56 & 534 & 30 \\
\hline
\end{tabular}
\end{small}
\end{table}

To assess the significance of differences across pipeline justifications, we conducted a Friedman test followed by Nemenyi post-hoc analysis \cite{nemenyi1963distribution}. The Friedman test revealed a highly significant difference across methods ($x^2 = 1823.83$, $p<0.0001$), rejecting the null hypothesis that all pipelines perform equally. 
Nemenyi post-hoc tests confirmed all pairwise differences exceed the critical difference (CD=0.064), \texttt{EM-RAG} vs \texttt{EM-KGRAG} ($|\bar{R}|=0.982>$ CD), \texttt{EM-RAG} vs \texttt{Baseline} ($|\bar{R}|=1.638>$ CD), and \texttt{EM-KGRAG} vs \texttt{Baseline} ($|\bar{R}|=0.656>$ CD).

\section{Conclusions and Future Work}
\label{sec:concl}
We propose \texttt{EmeraldMind}, a domain-specific RAG-based greenwashing detection framework comprising three variants. \texttt{EM-HYBRID} achieves the highest overall accuracy, combining \texttt{EM-RAG}'s broad coverage with \texttt{EM-KGRAG}'s accuracy. Few-shot prompting improves coverage without sacrificing accuracy, and domain-specific retrieval substantially enhances selective prediction without accuracy trade-offs. Future work should investigate how graph retrieval strategies (e.g., hop limits, schema constraints, traversal heuristics) influence performance and explore alternative hybrid methods for integrating textual and graph evidence. Incorporating additional knowledge sources, such as regulatory data and sustainability taxonomies, could further enhance coverage and justification quality.

\section*{Acknowledgment}

This work has been partially supported by project MIS 5154714 of the National Recovery and Resilience Plan Greece 2.0 funded by the European Union under the NextGenerationEU Program.
\bibliographystyle{ACM-Reference-Format}
\bibliography{bibliography}
\end{document}